\documentclass[10pt,twocolumn,letterpaper]{article}

\usepackage{wacv}              

\usepackage{graphicx}
\usepackage{amsmath}
\usepackage{amssymb}
\usepackage{booktabs}
\usepackage{enumitem}
\usepackage[accsupp]{axessibility}

\usepackage[pagebackref,breaklinks,colorlinks]{hyperref}

\usepackage[capitalize]{cleveref}
\usepackage{multirow}

\crefname{section}{Sec.}{Secs.}
\Crefname{section}{Section}{Sections}
\Crefname{table}{Table}{Tables}
\crefname{table}{Tab.}{Tabs.}

\def\wacvPaperID{1894} 
\def\confName{WACV}
\def\confYear{2024}

\begin{document}

\title{Bipartite Graph Diffusion Model for Human Interaction Generation}
\author{Baptiste Chopin$^1$, Hao Tang$^{2}$, Mohamed Daoudi$^{3,4}$\\
Inria, Université Côte d’Azur, France $^1$, \\ Carnegie Mellon University,
Pittsburgh, USA$^2$\\   Univ. Lille, CNRS, Centrale Lille, Institut Mines-Télécom, UMR 9189 CRIStAL, Lille, F-59000, France$^3$,\\ IMT Nord Europe, Institut Mines-Télécom, Univ. Lille, Centre for Digital Systems, Lille, F-59000, France$^4$}

\maketitle

\begin{abstract}
The generation of natural human motion interactions is a hot topic in computer vision and computer animation. It is a challenging task due to the diversity of possible human motion interactions. Diffusion models, which have already shown remarkable generative capabilities in other domains, are a good candidate for this task. In this paper, we introduce a novel bipartite graph diffusion method (BiGraphDiff) to generate human motion interactions between two persons. Specifically, bipartite node sets are constructed to model the inherent geometric constraints between skeleton nodes during interactions. The interaction graph diffusion model is transformer-based, combining some state-of-the-art motion methods. We show that the proposed  achieves new state-of-the-art results on leading benchmarks for the human interaction generation task. Code, pre-trained models and additional results are available at \url{https://github.com/CRISTAL-3DSAM/BiGraphDiff}.
\end{abstract}

\section{Introduction}

\label{sec:intro}
Modeling the dynamics of human motion is at the core of many applications in computer vision and computer graphics. However, the production of real human motion requires sophisticated equipment (e.g., expensive motion capture systems) and domain experts. In order to remove the skill requirements for users and potentially reach the general public, it is essential to create a human motion generation model capable of producing diverse motion sequences. Human motion generation is an active research area in robotics, computer graphics, and machine learning. It involves developing algorithms and techniques that enable machines to generate human-like motion for various applications such as animation, virtual reality, and robotics. Ongoing research efforts in this field aim to improve the quality, realism, and efficiency of motion generation methods, paving the way for innovative applications and breakthroughs in related fields. 

Several approaches have been proposed to predict or generate human motion for different tasks. Zhou \etal ~\cite{zhou2018autoconditioned} proposed an LSTM-based approach to generate long human motions. However the method suffers from error accumulation, a common problem with LSTMs. The authors also reported that their generative process sometimes converges toward a mean pose and stops generating motion. To avoid these issues some approaches used convolutional neural networks. Kosmas \etal~\cite{dance_gen_cnn} use CNNs to create a generative model for dance motions from music. However, using the samples generated by the method suffer from shaking caused by limited temporal coherency. 
More recently ACTOR~\cite{petrovich21actor}, a  Transformer-based variational autoencoder (VAE)  was designed to generate human motion based on the label. This approach limits the learned distribution to normal latent distribution since it mainly uses VAEs~\cite{KingmaICLR2014}. From this point of view, diffusion models are a better candidate for human motion generation, as they are free of assumptions about the target distribution. MotionDiffuse~\cite{zhang2022motiondiffuse} is a diffusion model-based text-driven motion generation framework. EDGE~\cite{TsengArxiv2022} uses a diffusion-based model that generates dance sequences conditioned on music. Tevet \etal ~\cite{TevetICLR2023} proposed a motion diffusion model (MDM) for generating human motion animation given an arbitrary condition or no condition. However, most works on human motion generation ignore human interactions and focus instead on the generation of motion of a single person. However, Human motion interaction modeling is a key component of video understanding and is indispensable because it is frequently observed in real video. It has various applications, such as a human-computer interface, sports, dance, game development, and an understanding of human behavior, and prediction. Generating interaction instead of single-person motion is also more challenging. Indeed, in addition to modeling the spatial and temporal of each skeleton for single-person motion generation, we need to model the interaction between the two skeletons. The motion of each skeleton influences the motion of the other. This is especially challenging since the interaction must be temporally coherent and require the motion of both persons to be synchronous if we want the interaction to be realistic.

In this paper, we address the challenge of generating high-quality 3D human motion interactions. The complexity lies in the nonlinear nature of human motion interactions and the diverse range of interactions between individuals. To tackle these challenges, we explore several key questions: how to effectively represent interactions between humans and how to model and generate diverse motion interactions. To address these questions, we propose~\textit{BiGraphDiff}, a graph Transformer denoising diffusion model that leverages the Transformer architecture. This approach enables us to generate diverse motion interactions while capturing the temporal dependencies of skeleton motion through the use of attention mechanisms. Additionally, the diffusion process incorporated in BiGraphDiff facilitates the generation of high-quality and diverse motions. In BiGraphDiff, we represent skeleton interactions using a bipartite graph~\cite{TangBMVC2020}. The bipartite graph serves as a means to capture the interactions between humans, with each human represented by a skeleton. By formulating the motion interaction generation as a reverse diffusion process, we can effectively generate realistic and diverse motion interactions. Overall, our contributions are summarized as follows:


\begin{itemize} [noitemsep,nolistsep, leftmargin=*]
    \item We propose the first Bipartite graph denoising diffusion model (BiGraphDiff) for human interaction generation. Our BiGraphDiff is able to generate motion interaction in a stochastic way, naturally leading to high diversity, and is able to generate very long motion sequences ($>$1000 frames).
    \item BiGraphDiff is a denoising diffusion process that learns not only the denoising of the motion but also it learns a Bipartite graph. The Bipartite graph aims to capture the relations between the two persons.
    

    \item BiGraphDiff achieves state-of-the-art quantitatively and qualitatively in action interaction and dance tasks. A user study shows that the generated sequences are better qualitatively than the sequences generated by state-of-the-art methods.
\end{itemize}


\section{Related Work}
\label{sec:relatedW}
We discuss the relevant literature from two perspectives, namely, previous methods of Human interaction motion synthesis and the literature on diffusion models.

\noindent \textbf{Human Interaction Motion Generation.}
Recently there has been an increase in motion generation based on different modalities, \cite{yin2021graphbased} use control signals such as the global trajectory of the person to generate human motion in long-term horizons while~\cite{ahuja-etal-2020-gestures} and~\cite{3dconvgesture_2021} generate motion based on speech audio. Meanwhile, others use only knowledge of the past motion which allows them to work in real-time but on shorter motion~\cite{martinez_human_2017,Cui_2020_CVPR,Sofianos_2021_ICCV}. More recently, several works have been dedicated  to human pose and motion generation from text or action labels, as well as its reciprocal task~\cite{GuoECCV22,Lucaseccv2022,chen2023executing,10096441}. These papers focus only on one person, while our approach is dedicated to the generation of two-person interactions. \cite{BaruahCVPRW2020} propose a multimodal variational recurrent neural network to predict the future motion of both participants in an interaction based on pasts sequences of motion. 
In contrast, we propose to generate human interaction between two persons.

\noindent \textbf{Generative Diffusion Models.}
Diffusion models~\cite{pmlr-v37-sohl-dickstein15,ho2020denoising} have shown great promise in terms of generative modeling by showing impressive results in synthesis applications ranging from image generation~\cite{DiffusionVideos2022}, audio-drive motion synthesis~\cite{alexanderson2022listen}, molecule generation~\cite{pmlr-v162-hoogeboom22a}, to text-driven motion generation \cite{saharia2022photorealistic}. More recently, some concurrent work in the field of text-to-motion introduces a diffusion-based method for generating text-conditioned motion.  
For example, Zhang~\etal~\cite{zhang2022motiondiffuse} propose MotionDiffuse, a diffusion model-based text-driven motion generation framework. Tseng~ \etal~\cite{TsengArxiv2022} propose EDGE, a method for generating editable dances that is able to create a realistic dance while remaining faithful to the original music. Dabral~\etal~\cite{dabral2022mofusion} introduce MoFusion, a denoising-diffusion-based framework for high-quality conditional human motion synthesis that can generate long and temporally plausible motions conditioned based on music or text. 
Despite achieving impressive performance, these methods use a diffusion-based method for generating the motion of only one person. In contrast, our proposed method BiGraphDiff proposes to generate the interaction between two persons and propose to learn a bipartite graph during the diffusion process. In addition, BiGraphDiff is applied for both text-to-motion and text-to-dance, and it is able to generate a long sequence of dance motions.

\section{Background}
\label{diff}
Denoising diffusion models (DDMs)~\cite{pmlr-v37-sohl-dickstein15} have emerged as powerful generative models, achieving outstanding results not only on image synthesis~\cite{ho2022imagen} but also for video synthesis \cite{singer2022make}. DDMs consist of two separate processes called forward diffusion and reverse diffusion. During the forward diffusion process, the data gradually is perturbed  by  Gaussian noise repeatedly until the data becomes Gaussian noise, while a neural model learns the reverse process of gradually denoising the sample.

Formally, the forward process on a real sample from a real data distribution $x_0{\sim} q(x)$ consists in a Markov chain that gradually adds noise following a variance schedule $\beta_t$ to obtain the posterior $q(x_{1:T}|x_0)$ with $x_1$ to $x_T$ the latent data:
\begin{equation}
\begin{aligned}
&q(x_{1:T}|x_0):= \prod_{t=1}^{T} q(x_t|x_{t-1}), \\
&q(x_t|x_{t-1}):=\mathcal{N}(x_t;\sqrt{1-\beta_t}x_{t-1},\beta_t\mathbf{I}).
\label{eq:Diffusion_forward}
\end{aligned}
\end{equation}




The reverse diffusion process, $p_\theta(x_{0:T})$, is a Markov chain that eliminates the noise from $x_T$ recursively until we obtain $x_1$. The reverse process can also be conditioned on an arbitrary condition $c$. With $p(x_T)=\mathcal{N}(x_T;\mathbf{0}, \mathbf{I})$:
\begin{equation}
\begin{aligned}
&p(x_{0:T}):= p(x_T)\prod_{t=1}^{T} p_\theta(x_{t-1}|x_t),\\
&p_\theta(x_{t-1}|x_t):=\mathcal{N}(x_{t-1};\mu_\theta(x_t,t,c), \Sigma_\theta(x_t,t,c)).
\label{eq:Diffusion_reverse}
\end{aligned}
\end{equation}

\begin{figure*}[t]
  \centering
   \includegraphics[width=1.0\linewidth]{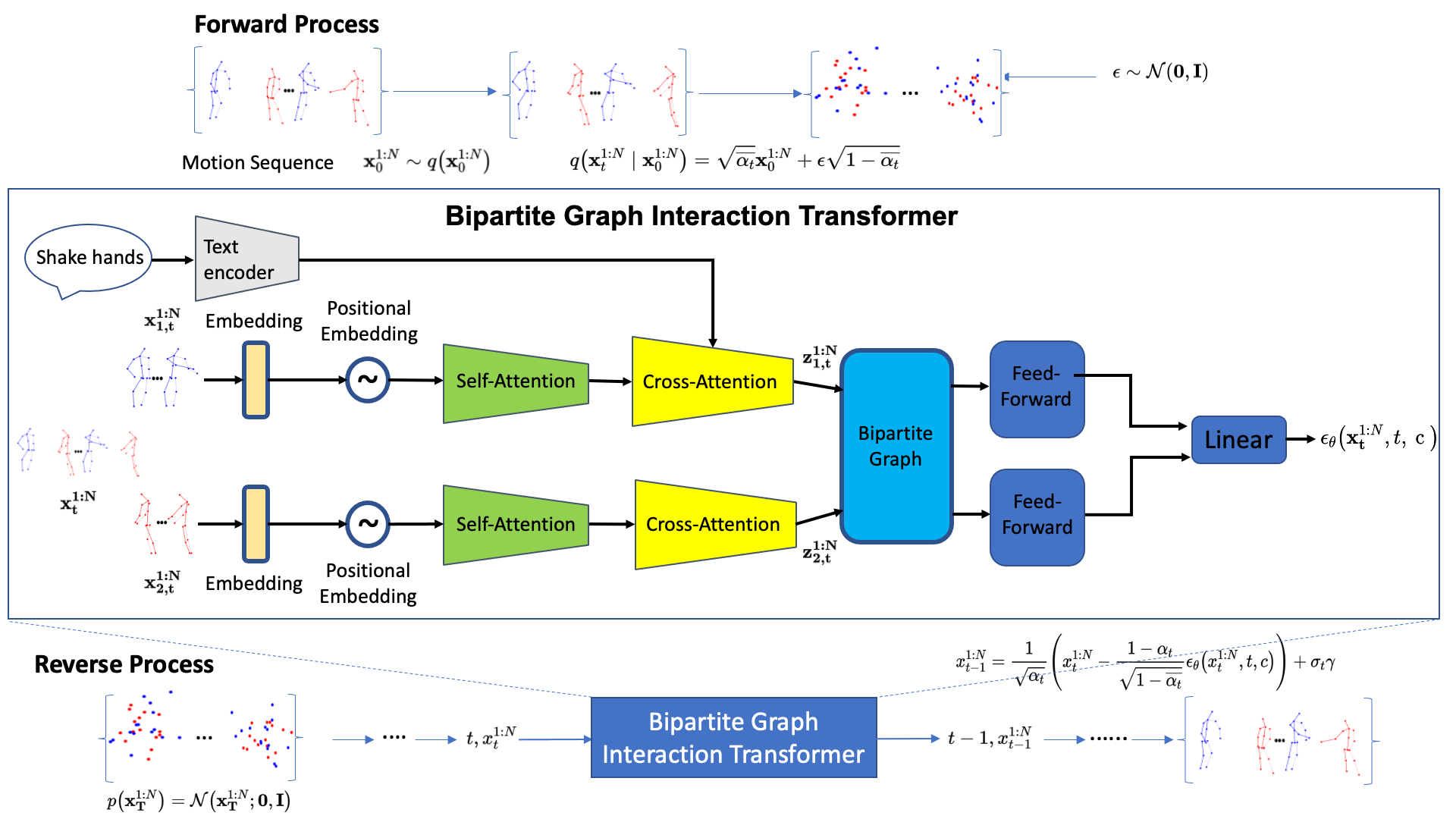}
   \caption{ \textbf{BiGraphDiff overview.} Top: the forward diffusion process to add noise to the motion sequence. Middle: the proposed Bipartite Graph Interaction Transformer to learn the denoising function. Bottom reverse diffusion process to generate motion sequence from noise.}
   \label{fig:overview}
   \vspace{-0.3cm}
\end{figure*}



During the denoising process the goal is to estimate $\mu_\theta(x_t,t,c)$ and $\Sigma_\theta(x_t,t,c)$. Based on ~\cite{Hao_diffusion} we can set $\Sigma_\theta(x_t,t,c){=}\sigma_t^2\mathbf{I}$ with $\sigma_t$ a constant and replace $\mu_\theta(x_t,t,c)$ as follow:

\begin{equation}
\mu_\theta(x_t,t,c)=\frac{1}{\sqrt{\alpha_t}}(x_t-\frac{1-\alpha_t}{\sqrt{1-\overline{\alpha_{t}}}}\epsilon_\theta(x_t,t,c)),
\label{eq:Diffusion_reverse_2}
\end{equation}
this means that we only need to estimate $\epsilon_\theta(x_t,t,c)$ to be able to denoise the latent data since we can recover $x_{t-1}$ using: 
\begin{equation}
x_{t-1}=\frac{1}{\sqrt{\alpha_t}}(x_t-\frac{1-\alpha_t}{\sqrt{1-\overline{\alpha_{t}}}}\epsilon_\theta(x_t,t,c))+\sigma_t\gamma,
\label{eq:Diffusion_reverse_3}
\end{equation}
with $\gamma{\sim}\mathcal{N}(\mathbf{0}, \mathbf{I})$. In our model we set $\sigma_t{=}\operatorname{log}(\beta_t\dfrac{1-\alpha_{t-1}}{1-\alpha_t})$ following \cite{Hao_diffusion} recommendation. To estimate $\epsilon_\theta(x_t,t,c)$, we will train a Bipartite Graph Interaction Transformer  (defined in section \ref{sec:BGIT}) to minimize the loss:
\begin{equation}
\begin{aligned}
L:=&E_{t\in[1,T],x_0\sim q(x_0),\epsilon\sim\mathcal{N}(\mathbf{0}, \mathbf{I})}[\|\epsilon-\epsilon_\theta(x_t,t,c)\|^2]\\
:=&E_{t\in[1,T],x_0\sim q(x_0),\epsilon\sim\mathcal{N}(\mathbf{0}, \mathbf{I})}[\|\epsilon-\epsilon_\theta(\sqrt{\overline{\alpha_{t}}}x_0\\&     +\epsilon\sqrt{1-\overline{\alpha_{t}}},t,c)\|^2].
\label{eq:Diffusion_loss}
\end{aligned}
\end{equation}

\section{Bipartite Graph Diffusion Model}
\label{sec:metho}
\subsection{Framework Overview}
As shown in Figure~\ref{fig:overview}, our goal is to generate a human motion interaction $x^{1:N}$ given an arbitrary condition $c$. Let us consider $x^{1:N}{=}\{x^1,\ldots, x^N\}$ an arbitrary sequence of joints that compose the two skeletons,  $x^i {\in} \mathbb{R}^{k \times 3 \times 2}$, where $k$ is the number of joints ($3$ the number of dimensions, $2$ the number of skeletons).  The motion generation is formulated as a reverse diffusion process that requires sampling a random noise $x_T^{1:N}$ from noise distribution to generate  a motion sequence as explained in section \ref{diff}. 
We propose Transformers to learn the denoising function and a bipartite graph to represent the relationship between the joints of the skeleton. The Bipartite Graph Interaction Transformer used by BiGraphDiff is based on the original Transformer~\cite{transformer}. It is composed of a text encoder, embedding and positional encoding layers, self-attention modules, cross-attention modules, a Bipartite graph module, feed-forward modules, and a final linear layer. 
The architecture is composed of a text encoder to encode the class label and a motion decoder to learn the interaction motion. This decoder deals with each skeleton separately with self-attention and cross-attention then our bipartite graph network learns a graph that represents the interaction and encodes this information in each skeleton. The proposed Transformer learns not only the denoising function but also the bipartite graph. 

\subsection{Bipartite Graph Interaction Transformer}
\label{sec:BGIT}

The text encoder is used to encode the class label $c$. We use a simple four layers Transformer encoder as described in ~\cite{transformer} that uses multi-head self-attention. To avoid training the encoder from scratch, we initialize the weight with those of CLIP~\cite{radford2021learning}. By using a text encoder instead of a simpler label encoder working with one hot vector, we allow our architecture to be directly used on more complex interaction datasets where motion is composed of several sub-motions when this kind of dataset will be available.
The motion decoder uses $x_t^{1:N}$ and the output of the text encoder to obtain $\epsilon_\theta(x_t,t,c)$. First we split $x_t^{1:N}$ into $x_{1,t}^{1:N}$ and $x_{2,t}^{1:N}$ which represent the first and second skeleton, respectively. Each skeleton passes through an embedding layer followed by a positional encoding layer introduced by \cite{transformer} that encodes the temporal information from each frame of the sequence. Then the data goes through self-attention and cross-attention layers. Attention is used to find correlations within the data. To reduce the complexity of attention layers, we use efficient attention \cite{shen2021efficient}. Each attention layer (self and cross) is followed by a stylization block. This module, introduced by \cite{zhang2022motiondiffuse}, allows the generative process to keep track of the current diffusion timestep $t$ improving the generation. The output of this module is added to the input of the attention through a residual connection. We then use a Bipartite graph module to learn the interaction between the two skeletons. After the bipartite graph module, the data of each skeleton goes through a feed-forward network. It is composed of linear projections, dropout, and GELU activation functions. It is followed by a stylization block to ensure that the information about the current timestep is not lost. The output is added to the input of the feed-forward network thanks to a residual connection. The Motion decoder contains eight identical layers and the input of layer $m$ is the output of layer $m-1$. Following those eight layers, the data of the two skeletons are concatenated and goes through a final linear projection to obtain $\epsilon_\theta(x_t,t,c)$ that we can use in our loss and to retrieve $x_{t-1}^{1:N}$.

\subsubsection{Bipartite Graph}
\begin{figure}[t]
  \centering
   \includegraphics[width=1.0\linewidth]{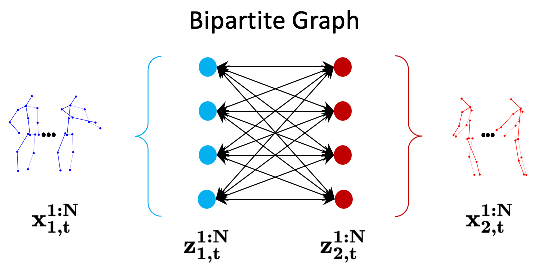}
   \caption{ \textbf{Illustration of the proposed bipartite graph method.}}
   \label{fig:bigraph}
   \vspace{-0.5cm}
\end{figure}

Here we describe in more detail the Bipartite graph module used in the motion decoder.
Following the self-attention and cross-attention module, we obtain embeddings for both skeletons, i.e., $z_{1,t}^{1:N}$ and $z_{2,t}^{1:N}$ (size $N\times d_l$, $d_l$ size of the embedding of a single skeleton). These embeddings go through the bipartite graph module.
The proposed bipartite graph aims to capture the long-range cross relations between the two embeddings $S_a {=} z_{1,t}^{1:N}$ and $S_b {=} z_{2,t}^{1:N}$ in a bipartite graph via GCNs.
Each node in $S_a$ is connected to all the nodes in $S_b$, as shown in Figure \ref{fig:bigraph}.

Firstly, $S_a$ and $S_b$ are separately fed into two encoders to obtain the feature $F_{a}$ and $F_{b}$, respectively.
We then reduce the dimension of $F_{a}$ with the function $\varphi_a(F_{a}) {\in} \mathbb{R}^{C \times D_a}$, where $C$ is the number of feature map channels, and $D_a$ is the number of nodes of $F_{a}$.
Meanwhile, we reduce the dimension of $F_{b}$ with the function $\theta_b(F_{b}) {=} H_b^\intercal {\in} \mathbb{R}^{D_b  \times C}$, where $D_b$ is the number of nodes of $F_{b}$.
Next, we project $F_{a}$ to a new feature $V_a$ in a bipartite graph using the projection function $H_b^T$. 
Thus we have:
\begin{equation}
V_a = H_b^\intercal \varphi_a(F_{a}) = \theta_b(F_{b}) \varphi_a(F_{a}),
\end{equation}
where both functions $\theta_b(\cdot)$ and $\varphi_a(\cdot)$ are implemented using a $1{\times1}$ convolutional layer. 
This results in a new feature $V_a {\in} \mathbb{R}^{D_b \times D_a}$ in the bipartite graph, which represents the cross relations between the nodes of the skeleton $F_{b}$ and the skeleton $F_{a}$.

After projection, we employ a fully connected bipartite graph with adjacency matrix $A_a {\in} \mathbb{R}^{D_b \times D_b}$.
We then use a graph convolution to learn the long-range cross relations between the nodes from both skeletons, which can be represented as:
\begin{equation}
\begin{aligned}
M_a = ({\rm I} - A_a) V_a W_a,
\end{aligned}
\end{equation}
where $W_a {\in} \mathbb{R}^{D_a \times D_a}$ denotes the trainable edge weights.
We use Laplacian smoothing  \cite{chen2019graph,li2018deeper} to propagate the node features over the bipartite graph. 
The identity matrix~${\rm I}$ can be viewed as a residual sum connection to alleviate optimization difficulties. 
We randomly initialize both the adjacency matrix $A_a$ and the weights $W_a$ and then train them by gradient descent.

After the cross-reasoning process, the new updated feature $M_a$ is mapped back to the original coordinate space for further processing.
Next, we add the result to the original feature $F_{a}$ to form a residual connection, as follows:
\begin{equation}
\begin{aligned}
\tilde{F}_{a} = \phi_a(H_b M_a) + F_{a},
\end{aligned}
\end{equation}
where we reuse the projection matrix $H_b$ and apply a linear projection
$\phi_a(\cdot)$ to project $M_a$ back to the original coordinate space. 
Therefore, we obtain the feature $\tilde{F}_{a}$, which has the same dimension as the original one $F_{a}$.

Similarly, we can obtain the new feature $\tilde{F}_{b}$. 
Overall, the proposed method reasons the cross relations between feature maps of different skeletons using a bipartite graph.

\begin{table}[!t] 
	\centering
		\caption{Classification score on NTU-26.}
  \resizebox{1.0\linewidth}{!}{%
\begin{tabular}{@{}c| c c c c c } 
\toprule
Method & GT  & ACTOR \cite{petrovich21actor} & MotionDiffuse \cite{zhang2022motiondiffuse} & BiGraphDiff \\
\midrule
\multicolumn{5}{c}{Classification Accuracy $\uparrow$}\\
Punching            & 76.0\%      & 1.0\%       & 43.0\%      & \textbf{49.0\% }     \\
Kicking             & 86.0\%      & 14.0\%      & 61.0\%      & \textbf{86.0\%}      \\
Pushing             & 97.0\%      & 77.0\%      & \textbf{86.0\%}      & 74.0\%      \\
Pat on back         & 88.0\%      & 4.0\%       & 72.0\%      & \textbf{80.0\%}      \\
Point Finger        & 83.0\%      & 0.0\%       & 52.0\%      & \textbf{76.0\%}      \\
Hugging             & 97.0\%      & 59.0\%      & 90.0\%      & \textbf{97.0\%}     \\  
Giving object       & 91.0\%      & 34.0\%      & 68.0\%      & \textbf{86.0\%}      \\ 
Touch pocket        & 93.0\%      & 35.0\%      & 81.0\%      & \textbf{84.0\%}      \\
Shaking hands       & 89.0\%      & 16.0\%      & 80.0\%      & \textbf{90.0\%}      \\
Walking toward      & 93.0\%      & 72.0\%      & 98.0\%      & \textbf{99.0\%}      \\
Walking apart       & 95.0\%      & \textbf{90.0\%}      & \textbf{90.0\%}      & \textbf{90.0\%} 	  \\
Hit with object     & 44.0\%      & 8.0\%       & 23.0\%      & \textbf{28.0\%}      \\ 
Wield knife         & 50.0\%      & 7.0\%       & 31.0\%      & \textbf{41.0\%}       \\ 
Knock over          & 85.0\%      & 4.0\%       & \textbf{61.0\%}      & \textbf{61.0\%}      \\
Grab stuff          & 74.0\%      & 0.0\%       & 57.0\%      & \textbf{62.0\%}      \\
Shoot with gun      & 57.0\%      & 1.0\%       & \textbf{46.0\%}      & 44.0\%      \\
Step on foot        & 89.0\%      & 5.0\%       & 85.0\%      & \textbf{90.0\%}      \\
High five           & 90.0\%      & 4.0\%       & 75.0\%      & \textbf{78.0\%}      \\
Cheers and drink    & 90.0\%      & 16.0\%      & 69.0\%      & \textbf{92.0\%}      \\
Carry object        & 96.0\%      & \textbf{98.0\%}      & 92.0\%      & 95.0\%      \\
Take a photo        & 87.0\%      & 19.0\%      & 63.0\%      & \textbf{80.0\%}      \\ 
Follow              & 94.0\%      & 68.0\%      & \textbf{90.0\%}      & 81.0\%      \\
Whisper             & 83.0\%      & 0.0\%       & 72.0\%      & \textbf{79.0\%}      \\
Exchange things     & 88.0\%      & 6.0\%       & 65.0\%      & \textbf{78.0\%}      \\
Support somebody    & 94.0\%      & \textbf{100.0\%}     & 94.0\%      & 92.0\%      \\
Rock paper scissor  & 91.0\%      & 6.0\%       & 75.0\%      & \textbf{91.0\%}      \\
Average             & 84.6\%      & 30.7\%      & 70.0\%      & \textbf{77.0\%}      \\
 \bottomrule
 \end{tabular}}
	\label{tab:quantitative}
 \vspace{-0.3cm}

\end{table}

\section{Experiments}

\subsection{Datasets}
There are few 3D motion two-person interaction datasets. 
Therefore we focus on two complementary datasets. The NTU RGB+D 120 dataset~\cite{NTU}, among its 120 classes, contains 26 classes labeled as ``Mutual Actions / Two Person Interactions'' which show two persons performing simple interaction motions. We take this 26-class subset that we call NTU-26 and split each class randomly to obtain our training and testing set. The testing set contains 2,600 samples (100 per class), and the training set is 19,787 samples. The framerate for NTU-26 is 30fps. The second dataset is DuetDance~\cite{duetdance}, which contains five classes of two-person dance motions for a total of 406 sequences. The motions are more complex than the one from NTU-26 and harder to classify, even for a human observer. The original dataset contains motions with great variations in lengths from 100 frames to more than 4,000. The average length is 483 frames with a  median of 360 frames. While our model can generate very long motions it causes a problem when obtaining quantitative results and lower the quality of the generation due to the limited presence of some sequence of certain lengths. We decided to split the sequences into subsequences of 300 frames or less. This increases the number of samples to train our network with. This increased number of samples will also help the diffusion model since it needs a lot of data. This leaves us with 698 training samples and 125 test samples (25 per class randomly selected). The skeletons from DuetDance are extracted from YouTube videos. The frame rate is either 25 fps or 30 fps, depending on the video. We use the same normalization for both datasets. First, we compute the means of the x,y, and z values of both skeletons. Then we subtract these means from the x,y, and z values. We divide the results by the Frobennius norm. Finally, we center the two skeletons around a point that is the middle of a segment that links the torso joints of both skeletons.

\subsection{Implementation Details}
We train our model on an NVIDIA A100 80Go GPU with PyTorch with a batch size of 128 for NTU and 64 for DuetDance. We train on NTU for 1,500 epochs and for 30,000 epochs on DuetDance.

\subsection{Baselines}
As there are currently limited interaction generation methods available, we compared our architecture to single-person motion generation methods. To ensure a fair comparison, we focused on generation methods conditioned on text or class labels and excluded methods conditioned on music, such as EDGE~\cite{TsengArxiv2022}. We compared our BiGraphDiff method to two state-of-the-art methods: MotionDiffuse~\cite{zhang2022motiondiffuse} and ACTOR~\cite{petrovich21actor}. MotionDiffuse is a recent architecture based on diffusion and transformers that generate single-person motion from the text. We used the code provided by the authors and ran the model with the recommended parameters. To ensure consistency, we used the same batch size and number of epochs as for our BiGraphDiff method. We chose MotionDiffuse over other diffusion methods because it has available code and its results are among the best, comparable to, or better than MoFusion~\cite{dabral2022mofusion}. ACTOR is a Transformer VAE method that also generates single-person motion. We used the code provided by the authors and retrained it on our datasets with recommended parameters. We had to deactivate the SMPL~\cite{SMPL:2015} loss function because it is not available in the NTU RGB+D and DuetDance datasets.  These two methods are mainly intended to generate the motion of a single individual. However, we can use them to generate the motion of two individuals by modifying their inputs. These methods assume that the motion is represented by a matrix with dimensions of $3k\times N$ ($k$ is the number of joints, and $N$ is the number of frames). Instead, we introduce a matrix twice as large, with dimensions of $6k\times N$, which consists of the concatenation of two skeletons. By doing so, neural networks observe an increase in the number of joints and can generate interacting motion.

\begin{table}[!t] 
	\centering
		\caption{FVD and Multimodality on NTU-26.}
		\resizebox{0.75\linewidth}{!}{%
\begin{tabular}{@{}c| c c } 
\toprule
Method                                      & FVD$\downarrow$       & Multimodality$\downarrow$\\
\midrule
ACTOR \cite{petrovich21actor}               & 25298.73          & 34.91   \\
MotionDiffuse \cite{zhang2022motiondiffuse} & 1292.32            & 14.94   \\ 
BiGraphDiff                                      & \textbf{1048.13}           & \textbf{11.28}  \\          
\bottomrule
\end{tabular}}
\label{tab:FVD}
\end{table}

\begin{figure*}[t]
  \centering
   \includegraphics[width=1.0\linewidth]{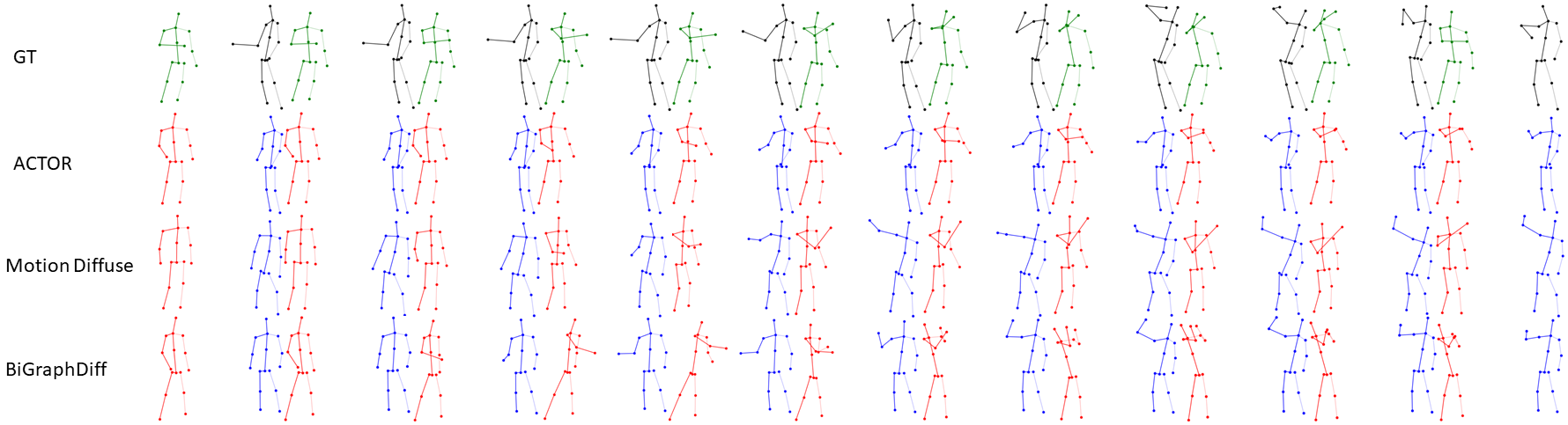}

   \caption{Examples of diverse motion generation for a given text prompt ``Cheers and Drink'' action from NTU.}
   \label{fig:cheers}
\end{figure*}

\vspace{-0.1cm}

\begin{table}[!t] 
	\centering
		\caption{Classification score on DuetDance.}
  \resizebox{1.0\linewidth}{!}{%
\begin{tabular}{@{}c| c c c c c } 
\toprule
Method & GT  & ACTOR \cite{petrovich21actor} & MotionDiffuse \cite{zhang2022motiondiffuse} & BiGraphDiff \\
\midrule
\multicolumn{5}{c}{Classification Accuracy $\uparrow$}\\
Cha-cha         & 28.0\%      & \textbf{36.0\%}       & 32.0\%      & 32.0\%      \\
Jive            & 52.0\%      & 16.0\%       & \textbf{20.0\%}      & 16.0\%      \\
Rumba           & 56.0\%      & 16.0\%       & 48.0\%      & \textbf{68.0}\%      \\
Salsa           & 88.0\%      & 0.0\%        & 64.0\%      & \textbf{76.0}\%      \\
Samba           & 52.0\%      & \textbf{80.0}\%       & 32.0\%      & 52.0\%      \\
Average         & 55.2\%      & 29.6\%       & 39.2\%      & \textbf{48.8}\%      \\
 \bottomrule
 \end{tabular}}
	\label{tab:quantitative_DD}

 \vspace{-0.4cm}

\end{table}

\subsection{Quantitative Results}
We perform the quantitative evaluation by using classification accuracy, Frechet Video Distance (FVD) score\cite{UnterthinerSKMM19}, and Multimodality. 

\noindent\textbf{Fréchet Video Distance (FVD)}  adapts the Fréchet Inception distance (FID) \cite{FID} for video sequences~\cite{UnterthinerIclrW2019}. FVD computes the distance between the generated data distribution and the ground truth using deep features.

\noindent\textbf{MultiModality.} 
Multimodality measures the diversity of generated samples in each class. It is defined as the average deep features distance of the samples generated by a method compared to the average deep features distance of the ground truth on a specific class. The deep features are extracted from the classifier used for classification accuracy To compute the average deep features distance we split the set of features of each class into two equal sets and compare the Euclidean norm between the pairs formed by a member of each set and compute the average over the size of the subsets. The Multimodality score is similar the multimodality from \cite{zhang2022motiondiffuse} but in our experimental results, we directly compare it with the multimodality of the ground truth.



\noindent\textbf{The classification accuracy} is obtained using a simple Transformer encoder followed by an MLP. The classifier is trained and tested on the same set as the generative methods.  \\

\noindent\textbf{NTU-26.} 
Table~\ref{tab:quantitative} shows that our method outperforms the two the-state-of-art methods in terms of average accuracy. BiGraphDiff outperforms MotionDiffuse by 7.0\% and ACTOR by 46.3\%. We are also very close to the accuracy of the classifier on the ground truth. This shows that the sequences generated by our method are realistic and correspond to the input class. In more detail, we can see that we outperform or equate the other methods on 22 classes out of 26. MotionDiffuse and ACTOR are both better in 2 classes. However, we can see that ACTOR results being actually better is debatable as some classes have very low accuracy, down to 0\%. We can also see that the classes in which we perform the worse (i.e., ``Hit with object'' 28\%, ``Wield knife'' 41\%, and ``Shoot with gun'' 44\%) are the ones where the results are also low for the ground truth. Those are classes where the main difference is the object used which is something we can not see using 3D skeleton data.
Table \ref{tab:FVD} shows the FVD and multimodality results. In terms of FVD and Multimodality, our method also outperforms the two other methods indicating that our method produces sequences closer to the real data. One issue when using the NTU dataset is that it is very noisy (see the ground truth in the qualitative results). This means that it is harder to generate noiseless sequences but also that a method that generates samples without noise might be disadvantaged in the quantitative results since they are compared with the ground truth and the classifier is trained on the noisy data.

\begin{table}[!t] 
	\centering
		\caption{FVD and Multimodality on DuetDance.}
	    \resizebox{0.75\linewidth}{!}{%
\begin{tabular}{@{}c| c c } 
\toprule
Method                                      & FVD$\downarrow$        & Multimodality$\downarrow$\\
\midrule
ACTOR \cite{petrovich21actor}               & 2641.08        & 67.79  \\
MotionDiffuse \cite{zhang2022motiondiffuse} & 1133.51          & 12.24\\ 
BiGraphDiff                                       & \textbf{997.92}           & \textbf{4.33}  \\          
\bottomrule
\end{tabular}}
\label{tab:FVD_DD}

\vspace{-0.4cm}

\end{table}

\noindent\textbf{DuetDance.}
Table \ref{tab:quantitative_DD} shows the classification results on the DuetDance dataset. We can see that, as on NTU-26, we have the best performance on average. The accuracy of our method is 9.6\% higher than on MotionDiffuse and only 6.4\% lower than on the ground truth. We can note that the accuracy for the ground truth is much lower than for NTU-26. This is due to the nature of the motion in DuetDance. The dance motions are much harder to recognize even for a human and also longer so it is not surprising that the results are worse. ACTOR, on the other only achieves results slightly higher than chance (20\%) we will see in the qualitative results that on DuetDance, ACTOR does not produce any motion, we will discuss this in detail in the qualitative results. In the ``jive'' class, all methods only achieve chance level or lower accuracy. But the results are not so low for the ground truth with means that all methods have trouble generating motion of the ``jive class''. In Table \ref{tab:FVD_DD}, we show that we outperform the other methods on both metrics meaning that the results of our method are more realistic.

\subsection{Qualitative Results}

\noindent\textbf{NTU-26.} 
Figure \ref{fig:cheers} shows visuals of sequences generated for the ``Cheers and drinks'' class. This class of motion is more complex than others because it is composed of two separate motions ``cheers'' and ``drink''. All methods generate a proper motion but ACTOR shows a low intensity for ``cheers'' and does not really generate the ``drink'' motion. MotionDiffuse generates a good motion with both ``cheers'' and ``drink'' but there is some noise and the arm length grows over time. Our method generates the proper motion with the two steps and does not produce the noise that is present in the ground truth. In our case, one character drinks while grabbing the glass with one hand while the other uses both hands showing the diversity in the generated motions. Overall we see that our motion is more realistic, temporally, and spatially coherent and manages well to keep the interaction coherent.

\begin{figure*}[t]
  \centering
   \includegraphics[width=1.0\linewidth]{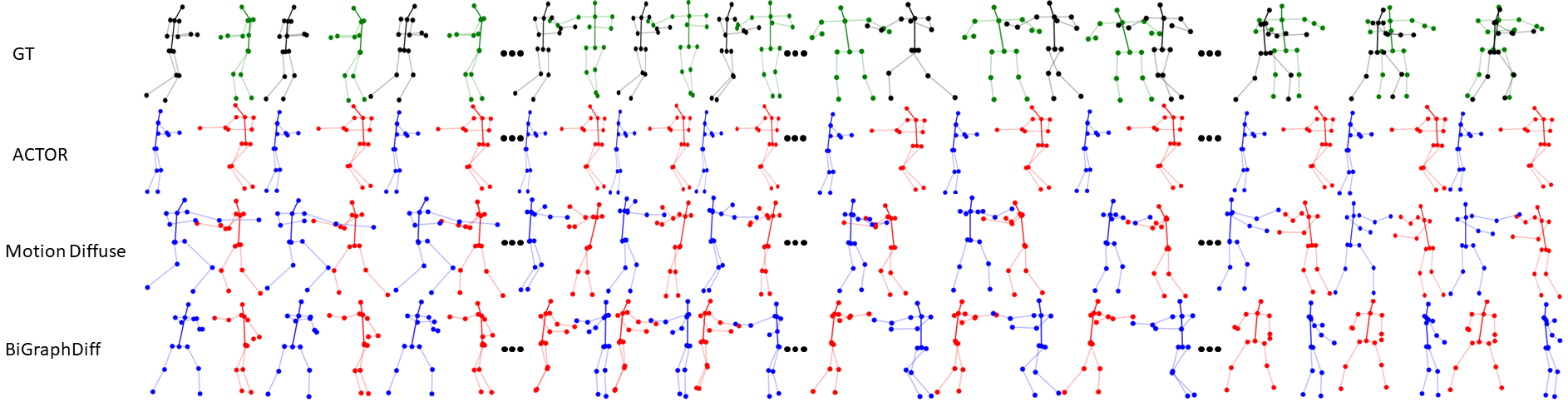}

   \caption{ Examples of diverse motion generation for a given text prompt ``Salsa'' action from DuetDance. }
   \label{fig:salsa}
\end{figure*}

\noindent\textbf{DuetDance.}
 Figure~\ref{fig:salsa} shows examples of  motion generation of the ``salsa'' class. The dance motions are more complex and the sequences are longer than NTU-26 sequences. ACTOR does not produce motion. We believe this to be due to the great variability of motions from the same class. ACTOR converges to a mean and finds that an unmoving pair of skeletons is the best generation for its losses. We see that MotionDiffuse produces a dance motion without noise. This is because there is less noise in DuetDance than in NTU-26. Our method also generates a dance motion but is better than MotionDiffuse, we reproduce the motion of characters changing sides that is present in the ground truth and the interaction is better as the arm of both characters does not overlap.

\subsection{User Study}
The user study compared BiGraphDiff with two leading methods (i.e., ACTOR \cite{petrovich21actor}, MotionDiffuse \cite{zhang2022motiondiffuse}) and the ground truth sequence. For both datasets, we randomly select 20 samples for each class from the test data. For each comparison, 30 participants are asked to answer two questions, i.e., `Q1: Which skeleton sequence is more realistic?', and `Q2: Which skeleton sequence matches the input text better?'. The numbers indicate the preference percentage of users who favor the results of the corresponding methods or the GT skeleton sequence. The results highlight the quality of the sequence generated by our method.


\begin{table}[!t] 
\caption{User study results ($\%$).}
	\centering
	\resizebox{0.75\linewidth}{!}{%
		\begin{tabular}{rcccc} \toprule
			\multirow{2}{*}{Method}  & \multicolumn{2}{c}{NTU-26} & \multicolumn{2}{c}{DuetDance}  \\ \cmidrule(lr){2-3} \cmidrule(lr){4-5}
			& Q1 & Q2 & Q1 & Q2   \\ \midrule
			ACTOR \cite{petrovich21actor}   & 6.1 & 7.8 & 5.6  & 5.9  \\
			MotionDiffuse \cite{zhang2022motiondiffuse} & 22.4 & 24.3 & 21.8 & 23.7          \\
		    BiGraphDiff  & \textbf{31.6} & \textbf{32.7} & \textbf{28.5}  & \textbf{30.1}                    \\ 
			GT   & 39.9 & 35.2 & 44.1  & 40.3 \\ \bottomrule
	    \end{tabular}}
	\label{tab:amt1}
 \vspace{-0.5cm}
\end{table}
\begin{figure*}[t]
  \centering
   \includegraphics[width=1.0\linewidth]{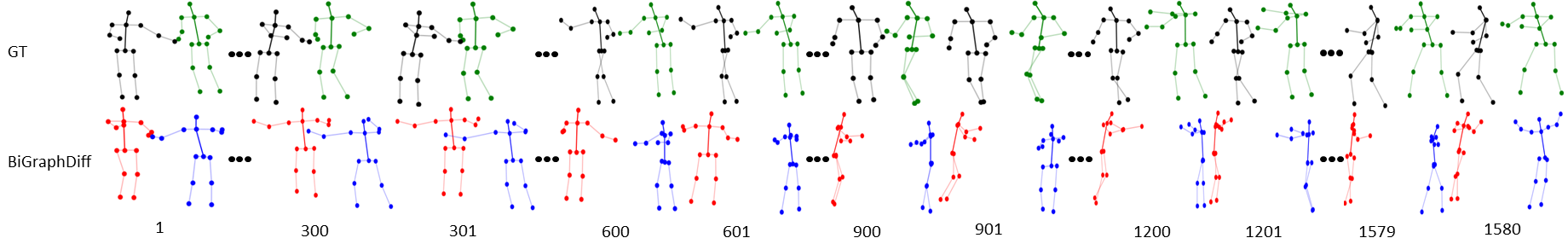}
   \caption{ Example of generation of very long sequences on the ``rumba'' class from DuetDance. Under each skeleton the frame number.}
   \label{fig:longgen}
   \vspace{-0.3cm}
\end{figure*}

\subsection{Ablation Study}
We report ablation
results in Table~\ref{tab:ablation} on the NTU-26 dataset. We compare a simple two-stream Transformer (S1), a two-stream Transformer in a diffusion process (S2), a two-stream Transformer in a diffusion process with a simple GCN (S3), and finally our method with bipartite graphs (S4). 

The results of S1 are extremely bad. It is explained by the fact the Transformer is a deterministic method and has a low generation diversity which explains the very high FVD. Furthermore, the noisy data from the NTU dataset makes it even harder to provide well-generated sequences. S2 provides much better results both in classification accuracy and FVD, the results are similar to the results obtained by MotionDiffuse. With S3 the simple GCN helps enhance the generation leading to better accuracy and FVD. This highlights the ability of the GCN to model more accurately the spatio-temporal dependencies from each skeleton. Adding a bipartite graph network in S4 provides a stronger increase in performance. It shows that modeling the interactions between the two skeletons is more important than trying to refine the interactions inside each skeleton as S3 did. It validates the use of the bipartite graph network in BiGraphDiff architecture.



\begin{table}[!t] 
	\centering
		\caption{Ablation study on NTU.}
		\resizebox{1\linewidth}{!}{%
\begin{tabular}{@{}l| c c } 
\toprule
Method                                      &Classification$\uparrow$   & FVD$\downarrow$   \\
\midrule

S1: Two Stream Transformer            &3.9\%   &21215.21       \\
S2: Two Stream Transformer + Diffusion                    &69.3\%   &1406.09     \\ 
S3: S2 + Simple GCN                   &73.2\%   &1123.88       \\ 
S4: S2 + Bipartite Graph              &\textbf{77.0\%}   &\textbf{1048.16}      \\      
\bottomrule
\end{tabular}}
\label{tab:ablation}

\vspace{-0.5cm}

\end{table}

\section{Very Long Generation}

Long-term motion generation
plays an important role in real-world applications. 
Our method is able to generate longer sequences, as shown in Figure~\ref{fig:longgen}. We train the network on the original DuetDance dataset with a maximum sequence length of 4050 frames. We use 376 samples for training and 40 (8 per class) for testing. Figure~\ref{fig:longgen} shows an example of 1580 frames from the ``rumba'' class. We can see that we generate dance-like motion for the entire duration of the sequence. However, it is very noticeable that we generate better motion for the first few hundred frames, we see that the motion quality around 300 frames is good, but then around 600 frames, we see deterioration that gradually becomes worse. This is due to the length of the sequences in the DuetDance dataset distribution which are usually not very long (average: 483 frames, median: 360 frames). Because of this, we do not use many very long positional encodings during training, preventing a good generation of very long motion.

\section{Conclusion, Limitations and Future Work}
We introduce the first approach for 3D  human motion interactions based on denoising diffusion models. Both quantitative and qualitative evaluations show that BiGraphDiff outperforms state-of-the-art methods. The proposed BiGraphDiff framework generates coherent human motion sequences that are longer and more diverse than the results of previous approaches. The proposed BiGraphDiff suffers however from the common limitations of diffusion models: the need for large datasets and the long training and testing duration. While the use of Bipartite graph improve performance by modeling the interaction, it also increase the complexity of the model. Our method has 221,476,320 trainable parameters while MotionDiffuse only has 86,990,688 and ACTOR 14,795,866. Recently modifications to the diffusion process have been investigated to improve the complexity of diffusion models~\cite{lu2022dpmsolver, bao2022estimating}. The method is also still slightly sensitive to noise in the training data and can sometimes generate deformed skeletons. This is due in part to the quality of the data used but also because we do not set any constraint related to the input data, e.g., bone length or relative position of joint for 3D skeletons. This means that BiGraphDiff can be used for tasks other than human interaction generation. As long as the input data can be split into two sets and has a temporal or positional component BiGraphDiff can be used for generation.

\section{Acknowledgments}
This project has received financial support from the CNRS through the 80—Prime program, from the French State, managed by the National Agency for Research (ANR) under the Investments for the future program with reference ANR- 16-IDEX-0004 ULNE.

\clearpage
\bibliography{ijcai23}


\clearpage

\section*{Supplementary material}

\subsection*{Additional training details}
The following values are the same for both NTU and DuetDance. We use train the network for 1000 diffusion steps with noise sampled in a linear manner and a learning rate of $0.0001$. The trained Bipartite Graph Interaction Transformer contains 8 identical layers and the multihead attention uses 8 attention heads. The Text encoder is a simple Transofoermer encoder with 4 layers and 4 attention heads.

\subsection*{Additional Qualitative Results}
\begin{figure*}[t]
  \centering
   \includegraphics[width=1.0\linewidth]{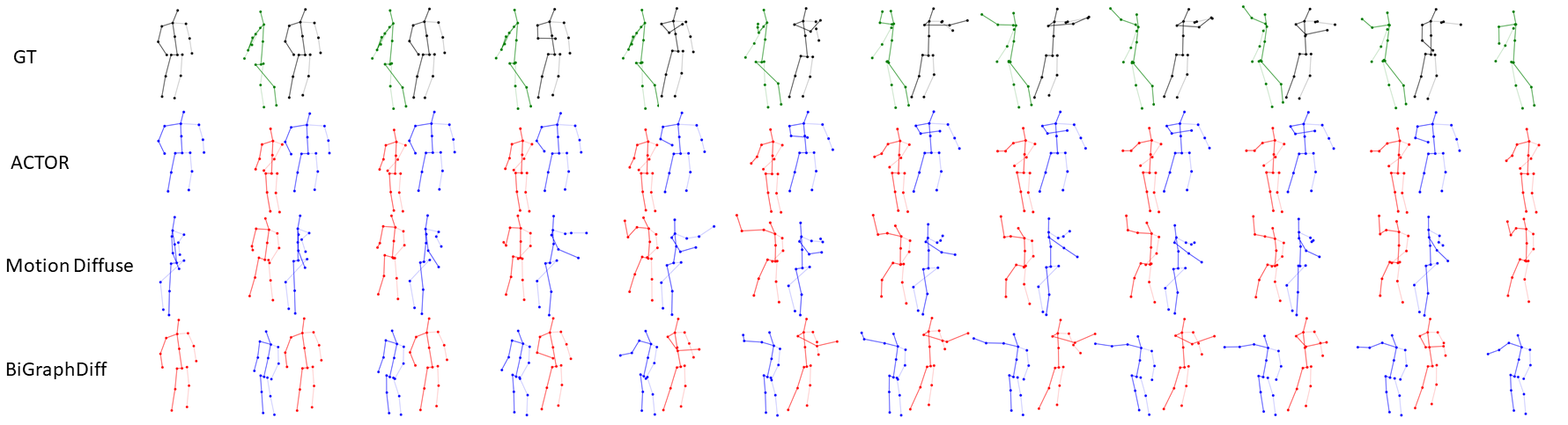}

   \caption{Examples of diverse motion generation for a given text prompt ``High-five'' action from NTU.}
   \label{fig:highfive}
\end{figure*}

\begin{figure*}[t]
  \centering
   \includegraphics[width=1.0\linewidth]{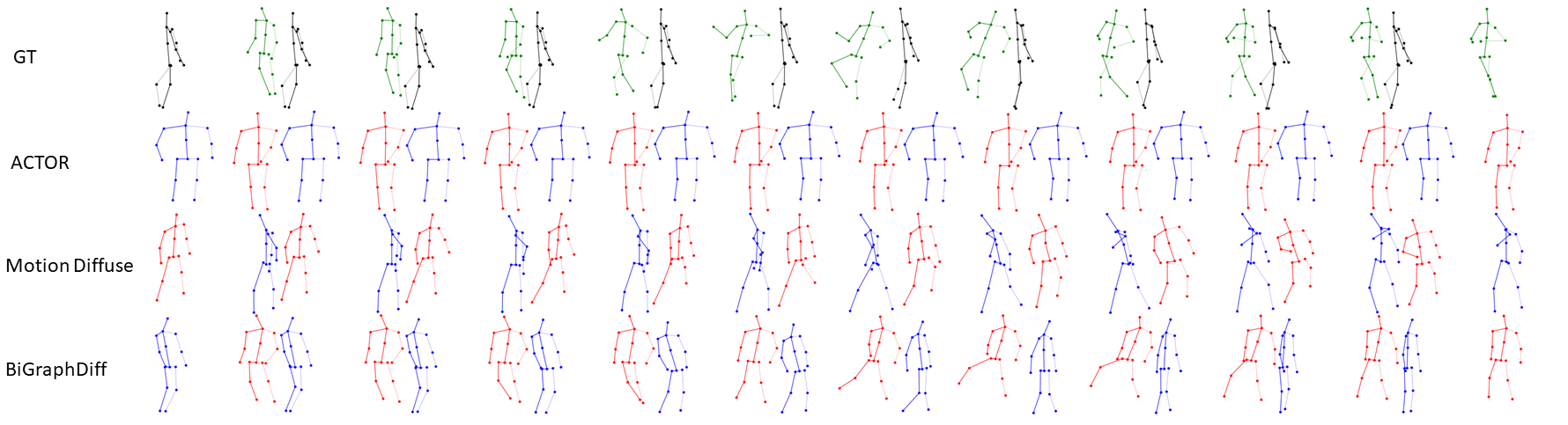}

   \caption{ Examples of diverse motion generation for a given text prompt ``Kicking'' action from NTU. }
   \label{fig:Kick}
\end{figure*}

\textbf{NTU-26.}
Figures \ref{fig:highfive} and \ref{fig:Kick} show visuals of sequences generated for the ``High-five'' and ``Kicking'' classes, respectively. For ``High-five'', ACTOR also generates a low-intensity motion, and both characters raise their hand but do not perform a high-five. Both MotionDiffuse and our method generate a high-five but MotionDiffuse shows noise and the hands of both characters stay far from each other. The ground truth once again contains noise that is not present in our generation. For the ``Kicking'' class, ACTOR does not generate any motion for either character. MotionDiffuse generates the red character as being kicked but does not generate the blue person kicking. On the other hand, our method generates both the kicking motion and the other character being kicked like the ground truth. In the ground truth, we can see that the leg is never fully extended during the kick. This is common for this class. The NTU-RGB+D dataset is captured using a Kinect camera and has difficulties capturing the legs due to the positioning of the camera and occlusion during interactions. This shows the kind of noise present in the original data again. Overall, we see that our motions are more realistic, temporally, and spatially coherent, and manage well to keep the interaction coherent.

\sloppy

\subsection*{Video Results}
Videos of the qualitative results presented in the main paper and this supplementary material can be found at \url{https://github.com/CRISTAL-3DSAM/BiGraphDiff} in examples directory. In ``Cheer\_and\_drink.mp4'', ``High\_five.mp4'', ``Kicking.mp4'', and ``salsa.mp4'' we show a comparison of the GT, the two state-of-the-art methods and BiGraphDiff on the four classes presented in the main paper and the supplementary material. In file ``very\_long\_rumba.mp4'', we show the video for very long-term generation on the rumba class presented in the main paper.

\subsection*{Additional animated results}
At \url{https://github.com/CRISTAL-3DSAM/BiGraphDiff} in `examples directory are additional animated results produced by BiGraphDiff. The visuals are in ``.gif'' format and can take some time to load for the longer sequences. ``BiGraphDiff\_NTU.zip'' contains the results on the NTU-26 dataset (100 samples per class) and ``BiGraphDiff\_DuetDance.zip'' contains the results on the DuetDance dataset (25 samples per class).

\subsection*{Code}
We provide the code and pretrained models at \url{https://github.com/CRISTAL-3DSAM/BiGraphDiff}. The code is for the training and testing of BiGraphDiff on NTU-26 and DuetDance, the quantitative evaluations of the generated sequence, and the visualization of the generated sequences. We provide the necessary formatted data to run the code for NTU. For DuetDance, the dataset is not public but can be shared upon request to the authors of \cite{duetdance}. We provide the code to format the raw data from \cite{duetdance} for use with BiGraphDiff. More details are provided in the README file provided with the code.

\end{document}